\documentclass[11pt,a4paper]{article}
\usepackage[hyperref]{acl2019}
\usepackage{times}
\usepackage{latexsym}

\usepackage{url}

\aclfinalcopy 

\title{Augmenting NLP data to counter Annotation Artifacts for NLI Tasks}

\author{Armaan Singh Bhullar \\
  \texttt{armaan.bhullar@utexas.edu} \\
  }

\date{November 2022}

\begin{document}
\maketitle
\begin{abstract}
  In this paper, we explore Annotation Artifacts - the phenomena wherein large pre-trained NLP models achieve high performance on benchmark datasets but do not actually "solve" the underlying task and instead rely on some dataset artifacts (same across train, validation and test sets) to figure out the right answer. We explore this phenomenon on the well-known Natural Language Inference task by first using contrast and adversarial examples to understand limitations to the model's performance and show one of the biases arising from annotation artifacts (the way training data was constructed by the annotators). We then propose a data augmentation technique to fix this bias and measure its effectiveness.
\end{abstract}

\section{Introduction}

Natural Language Inference (NLI), also known as Recognizing Textual Entailment (RTE), is the task of determining the inference relation between two (short, ordered) texts: entailment, contradiction, or neutral.  

In this paper, we study how Annotation Artifacts impact performance on an ELECTRA Small model \citep{clark2020electra} trained on the Stanford Natural Language Inference Dataset (SNLI) \citep{DBLP:journals/corr/BowmanAPM15}.  Next, we construct metrics to quantify the effects of this bias by identifying "easy" and "hard" examples on a subset of Validation data. Finally, we propose a solution using data augmentation during model training which counters this bias by forcing the model to learn "features" more relevant to the overall task.  We compare the results of this "modified" model with the "baseline" model and show improvement on both easy and hard examples.

\section{Model and Dataset}

We use the ELECTRA Small model trained using a GPU (NVIDIA 1060 Q Max) on the SNLI dataset whose train distribution can be seen in table \ref{snli-train} and test distribution in table \ref{snli-test}

\begin{table}[t!]
\begin{center}
\begin{tabular}{|l|l|}
\hline \textbf{Hypothesis Type} & \textbf{Number of Examples} \\ \hline
Entailment & 183416 \\
Neutral & 183187 \\
Contradiction & 183187  \\

\hline
\end{tabular}
\end{center}
\caption{\label{snli-train} SNLI Training dataset distribution}
\end{table}

\section{Training Baseline Model}
A model was trained using starter code provided from the repository\footnote{https://github.com/gregdurrett/fp-dataset-artifacts} on the training data, we call this the baseline model, performance metrics are available in table \ref{results-baseline}. 

\begin{table}[t!]
\begin{center}
\begin{tabular}{|l|l|l|l|}
\hline \textbf{Class} & \textbf{Precision} & \textbf{Recall (\%)} & \textbf{F1} \\ \hline
Entailment & 91 & 90 & 91 \\
Neutral & 85 & 86 & 86 \\
Contradiction & 91 & 90 & 91 \\
\textbf{Overall} & \textbf{89} & \textbf{89} & \textbf{89} \\

\hline
\end{tabular}
\end{center}
\caption{\label{results-baseline} Baseline model performance on Test set}
\end{table}

\section{Adversarial Attack}

\subsection{Setup and results}
We first explore the robustness of the model against adversarial attacks, here we follow the work of \citep{Wallace2019Triggers} and \citep{hossain-etal-2020-analysis} closely and use a similar style of adversarial attacks to prompt our model. We append a neutral sentence, " and false is not true." to the end of each example in the validation set and observe model performance, results are shown in this table\ref{results-baseline-adversarial}, clearly demonstrating that the model is unable to infer Entailments, instead mislabelling them as Neutral and Contradiction.

\begin{table}[t!]
\begin{center}
\begin{tabular}{|l|l|l|l|}
\hline \textbf{Class} & \textbf{Precision} & \textbf{Recall (\%)} & \textbf{F1} \\ \hline
Entailment & 79 & 7 & 12 \\
Neutral & 55 & 73 & 64 \\
Contradiction & 59 & 93 & 72 \\
\textbf{Overall} & \textbf{64} & \textbf{58} & \textbf{49} \\

\hline
\end{tabular}
\end{center}
\caption{\label{results-baseline-adversarial} Baseline model performance on adversarial examples - appending " and false is not true." to all examples in validation set}
\end{table}

We also look at some direct examples of the model's performance in table \ref{samples-baseline-adversarial}.

\subsection{Results Discussion - Explaining performance drop}

Looking at the classification report in table \ref{results-baseline-adversarial} and the confusion matrix in table \ref{confusion-matrix} we observe - 
\begin{itemize}
    \item Model is unable to recall Entailments, recall for Entailments drops from 90\% to 7\%
    \item Most of the Entailments actually get labeled as Neutral, one reason for this could be that Neutral is associated with more "unknown" words in hypothesis relative to premise while Entailment should have more words in common with the premise. This is known as Word Overlap Bias (WOB) and we will explore this further in the next sections
\end{itemize}

\begin{table}[t!]
\small
\begin{center}
\begin{tabular}{|p{1.75cm}|p{1.75cm}|p{1.25cm}|p{1.25cm}}
\hline \textbf{Premise} & \textbf{Hypothesis} & \textbf{Label} & \textbf{Predicted Labels} \\ \hline
The pets play in the apartment.	 & The animals are playing and false is not true.  & Entailment & Neutral \\
A person climbs a cliff face in the woods while others watch. & A person climbs a cliff face in the woods and false is not true.	 & Entailment & Neutral \\
Two men exchange conversation in a crowded place. & Two men exchange conversation in a crowded arena and false is not true.	 & Entailment & Neutral \\
This is a picture of three young men, dressed in suits with one on a bike, and of a young lady wearing a white dress.	& three guys in suit and false is not true. & Entailment & Contradiction \\
A greyhound jumps over a chain. & A dog ran away from the racetrack and false is not true.	 & Neutral & Contradiction \\

\hline
\end{tabular}
\end{center}
\caption{\label{samples-baseline-adversarial} Samples of Baseline model predictions under adversarial input}
\end{table}

\section{Proposing a solution - Data Augmentation}

One way to solve this is to modify the training data so it doesn't have such high Word Overlap Bias (WOB) following the approach of \citep{DBLP:journals/corr/abs-2005-04732}. To do this, we take neutral sentences and append them to the end of the sentences. We then let the model train keeping the rest of the parameters same.

\begin{table}[t!]
\begin{center}
\begin{tabular}{|l|l|}
\hline \textbf{Hypothesis Type} & \textbf{Number of Examples} \\ \hline
Entailment & 3368 \\
Neutral & 3237 \\
Contradiction & 3219  \\

\hline
\end{tabular}
\end{center}
\caption{\label{snli-test} Test data Distribution }
\end{table}

\begin{table}[t!]
\begin{center}
\begin{tabular}{|l|l|l|l|}
\hline \textbf{Predicted/Actual} & \textbf{Entailment} & \textbf{Neutral} & \textbf{Contradiction}\\ \hline
\textbf{Entailment} & 220 & 1803 & 1306 \\
\textbf{Neutral} & 31 & 2398 & 806 \\
\textbf{Contradiction} & 28 & 192 & 3058  \\

\hline
\end{tabular}
\end{center}
\caption{\label{confusion-matrix} Confusion matrix for Adversarial attack on Baseline model}
\end{table}

\subsection{Test metric to measure WOB bias}
We now propose a measure of how much WOB impacts model performance. Let's focus on entailment examples since they particularly rely on hypothesis and premise having common words and are likley to suffer the most from WOB, as demonstrated in the previous sections. To measure the WOB's effect on model performance, we calculate the percentage of tokens a hypothesis has in common with it's premise after tokenizing and lemmatizing, example calculations are shown in table \ref{jaccard} Using this as a measure of Word Overlap (WO), we then sort the entailment examples in Validation set by this metric and identify 1000 each having the highest and lowest WO. We call the examples with the highest WO as "Easy Entailment" and the lowest "Tough Entailment".  We then measure the performance of the baseline model on this set, shown in table \ref{metrics-1}.
\begin{table}[t!]
\begin{center}
\begin{tabular}{|l|l|l|}
\hline \textbf{Example Group} & \textbf{Model} & \textbf{Accuracy (\%)}\\ \hline
Easy Entailment & Baseline & 95.1 \\
Tough Entailment & Baseline & 86.2 \\

\hline
\end{tabular}
\end{center}
\caption{\label{metrics-1} Baseline model performance on Entailment set splits}
\end{table}

\begin{table}[t!]
\small
\begin{center}
\begin{tabular}{|p{1.25cm}|p{1.25cm}|p{1.25cm}|p{1.25cm}|p{1cm}}
\hline \textbf{Premise} & \textbf{Premise tokenized} & \textbf{Hypothesis} & \textbf{Hypothesis tokenized} & \textbf{WO}\\ \hline
Man wearing dark clothes walks down a street with a wall of graffiti to his left. & [man, wear, dark, cloth, walk, down, a, street, with, a, wall, of, graffiti, to, his, left.] & The man is painting the street.	& [the, man, is, paint, the, street.] & 0.2 \\
Two children are diving side by side into a river. & [two, children, are, dive, side, by, side, into, a, river.] & Two children are diving off a high rock into a small river. & [two, children, are, dive, off, a, high, rock, into, a, small, river.] & 0.64 \\
A small girl in a blue shirt with a posy pattern brushes her teeth with a red and yellow toothbrush. & [a, small, girl, in, a, blue, shirt, with, a, posi, pattern, brush, her, teeth, with, a, red, and, yellow, toothbrush.]	 & She punches through the wall & [she, punch, through, the, wall] & 0.0 \\

\hline
\end{tabular}
\end{center}
\caption{\label{jaccard} Example calculations for Word Overlap. NLTK is used for tokenizing and lemmatizing the sentences. WO equals the proportion of tokens in Hypothesis which are present in the Premise}
\end{table}

Here we can observe that the model misclassifies 9\% more entailment examples in cases where the WO is lower, thus our metric does indeed capture the impact of WOB on the model.

\subsection{Training an augmented model}

We retrain the ELECTRA model from scratch on a new set of training examples constructed by selecting one of the 6 sentences shown in table \ref{neutral-sentence} randomly and appending it to each of the training examples. The particular sentence to be appended is selected randomly and modifications made using the Dataset API of huggingface. 

\begin{table}[t!]
\begin{center}
\begin{tabular}{|l|}
\hline \textbf{Neutral Sentence} \\ \hline
 and false is no true \\ 
 and any true is true \\ 
 and false is never true \\ 
 and anything true is true \\ 
 and false is not true \\
\hline
\end{tabular}
\end{center}
\caption{\label{neutral-sentence} Neutral sentences used for augmenting training data}
\end{table}

\subsection{Results and Discussion}
2 comparisons are made with the baseline model. Table \ref{results-baseline} and \ref{results-augmented} compare model performance on the overall test set. Table\ref{results-2} compares performance on the 2 Entailment sets - easy and hard as described in previous section.   
As we can see, our augmented model has comparable overall performance, with a slightly higher recall observed for Entailment at the slight cost of precision for Neutral and Contradiction classes. On the two splits of Entailment examples, however, we observe that the modified model \textbf{outperforms the baseline by a small margin of 0.5\% to 0.7\%}. This lift of 0.7\% on the Hard entailment examples can be attributed to our data augmentation technique which forced the model to reduce it's dependence on WOB for identifying entailments, thus improving generalization ability.

\begin{table}[t!]
\begin{center}
\begin{tabular}{|l|l|l|l|}
\hline \textbf{Class} & \textbf{Precision} & \textbf{Recall (\%)} & \textbf{F1} \\ \hline
Entailment & 91 & 91 & 91 \\
Neutral & 84 & 86 & 85 \\
Contradiction & 92 & 89 & 90 \\
\textbf{Overall} & \textbf{89} & \textbf{89} & \textbf{89} \\

\hline
\end{tabular}
\end{center}
\caption{\label{results-augmented} Augmented model performance on Test set}
\end{table}

\begin{table}[t!]
\begin{center}
\begin{tabular}{|l|l|l|}
\hline \textbf{Example Group} & \textbf{Model} & \textbf{Accuracy (\%)}\\ \hline
Easy Entailment & Baseline & 95.1 \\
Easy Entailment & \textbf{Augmented} & \textbf{95.6} \\
Tough Entailment & Baseline & 86.2 \\
Tough Entailment & \textbf{Augmented} & \textbf{86.9} \\

\hline
\end{tabular}
\end{center}
\caption{\label{results-2} Augmented vs Baseline model comparison on Entailment set splits}
\end{table}

\section{Conclusion}
NLP research has accelerated with newer and larger Deep Learning based models releasing everyday and establishing SOTAs on a number of datasets, in this paper we examine the issue of if a model is really "learning" the underlying task instead of learning some biases which may creep in the datasets during annotation (annotation artifacts). We outline a way to measure this bias on a set of examples and understand how data augmentation can help improve the models real learning. We then show how, by incorporating this data augmentation we can improve the model's performance on both hard and easy examples for a similar overall performance.

\section*{Acknowledgments}

I thank the TAs and Prof. Gregg Durett of the University of Texas at Austin for their excellent guidance and help in the NLP course and for providing an opportunity to explore this very active and interesting research area in NLP.

\bibliography{acl2019}

\begin{thebibliography}{5}
\expandafter\ifx\csname natexlab\endcsname\relax\def\natexlab#1{#1}\fi

\bibitem[{Bowman et~al.(2015)Bowman, Angeli, Potts, and
  Manning}]{DBLP:journals/corr/BowmanAPM15}
Samuel~R. Bowman, Gabor Angeli, Christopher Potts, and Christopher~D. Manning.
  2015.
\newblock \href {http://arxiv.org/abs/1508.05326} {A large annotated corpus for
  learning natural language inference}.
\newblock \emph{CoRR}, abs/1508.05326.

\bibitem[{Clark et~al.(2020)Clark, Luong, Le, and Manning}]{clark2020electra}
Kevin Clark, Minh-Thang Luong, Quoc~V. Le, and Christopher~D. Manning. 2020.
\newblock \href {https://openreview.net/pdf?id=r1xMH1BtvB} {{ELECTRA}:
  Pre-training text encoders as discriminators rather than generators}.
\newblock In \emph{ICLR}.

\bibitem[{Hossain et~al.(2020)Hossain, Kovatchev, Dutta, Kao, Wei, and
  Blanco}]{hossain-etal-2020-analysis}
Md~Mosharaf Hossain, Venelin Kovatchev, Pranoy Dutta, Tiffany Kao, Elizabeth
  Wei, and Eduardo Blanco. 2020.
\newblock \href {https://doi.org/10.18653/v1/2020.emnlp-main.732} {An analysis
  of natural language inference benchmarks through the lens of negation}.
\newblock In \emph{Proceedings of the 2020 Conference on Empirical Methods in
  Natural Language Processing (EMNLP)}, pages 9106--9118, Online. Association
  for Computational Linguistics.

\bibitem[{Wallace et~al.(2019)Wallace, Feng, Kandpal, Gardner, and
  Singh}]{Wallace2019Triggers}
Eric Wallace, Shi Feng, Nikhil Kandpal, Matt Gardner, and Sameer Singh. 2019.
\newblock Universal adversarial triggers for attacking and analyzing {NLP}.
\newblock In \emph{Empirical Methods in Natural Language Processing}.

\bibitem[{Zhou and Bansal(2020)}]{DBLP:journals/corr/abs-2005-04732}
Xiang Zhou and Mohit Bansal. 2020.
\newblock \href {http://arxiv.org/abs/2005.04732} {Towards robustifying {NLI}
  models against lexical dataset biases}.
\newblock \emph{CoRR}, abs/2005.04732.

\end{thebibliography}
\bibliographystyle{acl_natbib}

\end{document}